# A Cognitive Memory Network

A. P. James and S. Dimitrijev

*A resistive memory network that has no crossover wiring is proposed to overcome the hardware limitations to size and functional complexity that is associated with conventional analog neural networks. The proposed memory network is based on simple network cells that are arranged in a hierarchical modular architecture. Cognitive functionality of this network is demonstrated by an example of character recognition. The network is trained by an evolutionary process to completely recognize characters deformed by random noise, rotation, scaling, and shifting.*

*Introduction:* Analog neural network hardware has many advantages over its digital and software counterparts. However, the existing network architectures for analog hardware implementations are limited by wiring complexity associated with the large number of crossover wirings resulting from its synaptic connections [1-3].
In this Letter, we present the concept of a hardware-based cognitive memory network with architecture that has no crossover wiring. The cognitive functionality of the network is demonstrated through an example of deformed character recognition.

*Network Cell and Architecture:* Fig. 1*a* shows a network cell that can be used in the network architecture shown in Fig. 1*b*. The input to the network are analog or digital voltages labeled by $V_i$ ($i = 1, 2, ..N$). Parameters of the cell are the resistors $R_i$ ($i = 1, 2..N$) and $R_o$, whereas the threshold voltage ($V_T$) of the inverter is kept constant for consistency with the standard CMOS technology. The resistors $R_i$ and $R_o$ are memory elements (e.g. gate controlled Flash [4], MRAM [5], QsRAM [6] etc) to enable the network to learn the desired function for a given task. The inverter is used to provide binary outputs from the cells.

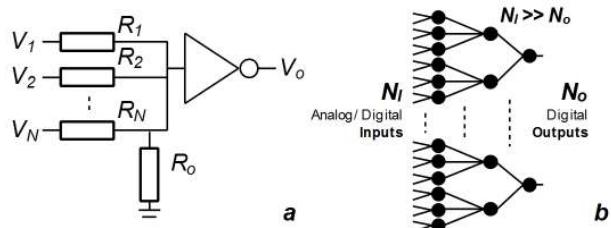

Fig. 1 Memory network cell and architecture.
*a* The cell structure.
*b* The network architecture (each node represents a cell) with no crossover wiring. The cells are arranged in a hierarchical manner with $N_I >> N_O$.

The network architecture having no crossover wiring is formed by many cells across several layers arranged in a hierarchical manner (Fig. 1*b*). The system architecture enables space-dimensionality reduction, as illustrated by the larger number of inputs ($N_I$) and the smaller number of digital outputs ($N_O$). The network is highly modular in terms of the arrangement of the cells between layers. Each cell can have different number of input resistors ($R_i$) within a layer or between the layers. This means that there can be many different ways of forming the network for the same given number of inputs and outputs. The parameters of the network are the complete set of resistor values used in the network, which are labeled as *Rset*. The inverters in the cell provide signal regeneration along the layers so that the network becomes almost immune to signal noise.

*Character recognition example:* The aim of this example is to illustrate that by using an evolutionary process the parameters of the memory network can be trained to recognize printed characters with random deformations. Each character in Fig. 2*a* is represented as a matrix of 36×36 pixels in digital values, so that the total number of required inputs to the network at a time becomes 1296. This network has 3 layers with 6, 6, and 3 inputs per cell in the first, second and third layers, respectively. The network is set with 12 outputs, which means that 12 bit codes are used to classify the characters. There are $2^{12}$ possible output codes, which allow more than one code to be assigned to each character. Although the network could function with a single code per character, the multiple-code approach can help to achieve better noise immunity. In this specific example, each character is assigned a main code and then the twelve codes that differ by one bit from the main code are added to form the set of codes for each character.
The evolutionary process used to setup the values of the network resistors (*Rset*) consists of selection, genetic, and training stages, as illustrated in Fig. 2*b*. These stages can be treated as a 3 level training algorithm that can be implemented as peripheral circuits on the same chip within the network. The average bit error per character (ABE), which is the net average of all the output bit errors, is used as a measure for ranking the *Rset*'s generated at each level.

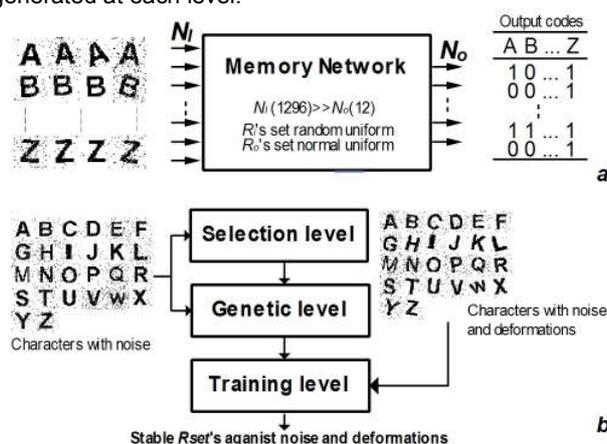

Fig. 2 Character recognition using memory network
*a* Illustration of the memory network used for recognition of characters with random noise, rotation, shift, and scaling.
*b* Evolutionary process used to set the parameters of the memory network (*Rset*'s) that provide the character-recognition functionality.

At the selection level, *Rset*'s are set to random values and the 26 characters are applied one by one to the inputs of the network (Fig. 2*b*). The first 400 *Rset*'s that give unique binary output codes for each of the character are selected. Among these 400 *Rset*'s, the five *Rset*'s with the lowest ABE that are processed against 12% standard salt and pepper noise are selected to be used in the genetic level.
The genetic level is designed to accelerate the search for stable *Rset* combinations against the effects of noise by using a genetic algorithm with two random crossover points without mutation. At the genetic level, 800 second generation of *Rset*'s are created and five of those with the lowest ABE are selected for the training level.

Training phase consists of 26 capital printed characters with the following deformations: random (salt and pepper) noise, clockwise and anticlockwise rotation, scaling, and horizontal and vertical shifts (Fig. 2*b*). These deformations are randomly generated following a normal distribution with standard deviations ($\sigma$) and a range of $\pm 3\sigma$ with $\sigma$ values of 4% for random noise, $5°$ for clockwise and anticlockwise rotation, 5% for scaling, and 5 pixels for horizontal and vertical shifts. The individual $R_i$'s that correspond to the outputs with the maximum bit errors are updated randomly until the network becomes immune to the character deformations. The resulting *Rset*'s provide a minimum binary spacing of 3 between the 26 main character codes. With this, the output codes that differ by 1 bit from the main codes can also be used for unambiguous recognition of the relevant character.

To test the performance of the memory network, we generated and applied $10^4$ test character sets with larger deformations compared to the deformation used during the training phase: $\sigma$ values of 12% for random noise, $\pm 15\%$ for scaling, $\pm 15°$ for rotation, and $\pm 15$ pixels for horizontal and vertical shifts. ABE of less than 0.5 was achieved with maximum bit error of 1 and as a result, testing with $26 \times 10^4$ deformed characters resulted in 100% recognition accuracy.

*Discussion:* A conventional analog hardware neural network, irrespective of its size, will have many crossover wirings. For example, a 3 layer network with 5 nodes per layer will have at least 6530 crossover wirings due to the connections to its weights (given that $2^5=32$, 5 nodes are absolutely necessary to be able to code 26 characters). The demonstrated memory network with the architecture shown in Fig 1*b* does not have any crossover wirings within or across the layers. As a consequence, the nodes in the output layer of the memory network are independent from each other and each output node defines a group of weights associated with it across the layers. The character recognition example demonstrates that highly localized training is possible and that the interconnections between the nodes, which are the basic feature of neural networks, are not necessary. The advantage of a memory network without crossover wiring and a highly modular structure is that it enables large network architectures that are impossible to implement and train with conventional neural networks.

Increasing the memory network size can enable application of the character recognition principle to other tasks, including face and speech recognition. These can be achieved by increasing the number of layers and inputs either in series or by replicating the network in parallel. The memory network can be integrated into camera systems for a broad range of applications. Offline applications require a limited number of programming cycles, which means flash memory can be used as the memory resistors, along with CMOS inverters integrated on a single chip. For real-time applications, such as object motion detection, it would require memories that can do large number of cycles and exhibit long data retention time which may be possible with MRAM [5] or QsRAM [6]. MRAM is a mature technology but has limited capacity, however higher capacity could be achieved in hardware with emerging memory technologies, such as QsRAM. It is also worth mentioning that faulty memory elements in the memory network appear as input noise to that cell, however, a properly trained network is stable to the effects of input noise. Because of this and because of the fact that the majority of the memory elements appear in the initial layers of the network, the proposed cognitive memory networks also provide a degree of robustness to yield problems.

*Conclusion:* This Letter demonstrated that cognitive memory networks with simple network cells and no crossover wiring, which removes a significant limitation of the conventional networks, can perform cognitive functions. The potential for increasing the system and functional complexity using the modular architecture of the memory network is enormous as the modern memory technologies can provide in excess of $10^9$ memory elements on a single chip.

*Acknowledgements:* Funding support by Qs Semiconductors is gratefully acknowledged.

Authors' affiliations:
A. P. James and S. Dimitrijev (Queensland Micro- and Nanotechnology Centre, Griffith University, Qld. 4111)

Emails: a.james@griffith.edu.au